\documentclass[10pt,twocolumn,letterpaper]{article}

\usepackage{wacv}              

\usepackage{multirow, makecell}
\usepackage{graphicx}
\usepackage{float}
\usepackage[accsupp]{axessibility}  

\newcommand{\best}[1]{\textbf{{#1}}}
\newcommand{\second}[1]{{#1}}
\definecolor{wacvblue}{rgb}{0.21,0.49,0.74}
\usepackage[pagebackref,breaklinks,colorlinks,allcolors=wacvblue]{hyperref}

\def\wacvPaperID{1830} 
\def\confName{WACV}
\def\confYear{2026}

\title{View-aware Cross-modal Distillation for Multi-view Action Recognition}

\author{
Trung Thanh Nguyen$^{1,2}$\thanks{Corresponding author: \texttt{nguyent@cs.is.i.nagoya-u.ac.jp}.}\hspace{5pt}, Yasutomo Kawanishi$^{2,1}$, Vijay John$^{3}$, 
Takahiro Komamizu$^{1}$, Ichiro Ide$^{1}$\\
$^{1}$Nagoya University, Japan \hspace{5pt}
$^{2}$Guardian Robot Project, R-IH, RIKEN, Japan\\
$^{3}$Lawrence Technological University, USA\\
}

\begin{document}

\maketitle
\begin{abstract}

The widespread use of multi-sensor systems has increased research in multi-view action recognition. While existing approaches in multi-view setups with fully overlapping sensors benefit from consistent view coverage, partially overlapping settings where actions are visible in only a subset of views remain underexplored. This challenge becomes more severe in real-world scenarios, as many systems provide only limited input modalities and rely on sequence-level annotations instead of dense frame-level labels. In this study, we propose \textbf{Vi}ew-aware \textbf{C}r\textbf{o}ss-modal \textbf{K}nowledge \textbf{D}istillation (ViCoKD), a framework that distills knowledge from a fully supervised multi-modal teacher to a modality- and annotation-limited student. ViCoKD employs a cross-modal adapter with cross-modal attention, allowing the student to exploit multi-modal correlations while operating with incomplete modalities. Moreover, we propose a View-aware Consistency module to address view misalignment, where the same action may appear differently or only partially across viewpoints. It enforces prediction alignment when the action is co-visible across views, guided by human-detection masks and confidence-weighted Jensen–Shannon divergence between their predicted class distributions. Experiments on the real-world MultiSensor-Home dataset show that ViCoKD consistently outperforms competitive distillation methods across multiple backbones and environments, delivering significant gains and surpassing the teacher model under limited conditions.

\end{abstract}

\section{Introduction}
\label{sec:introduction}

The increasing use of multi-sensor systems in smart homes~\cite{nguyen2025multisensor}, surveillance~\cite{li2022attention}, and assistive robotics~\cite{abdelaal2020multi} has driven growing interest in multi-view action recognition. 
By aggregating observations from multiple viewpoints, such systems capture richer spatio-temporal cues than single-view setups and improve robustness to occlusion, viewpoint changes, and background clutter~\cite{wang2014cross, shahroudy2016ntu, liu2019ntu, olagoke2020literature}.
Most existing methods, however, are designed for fully overlapping sensor setups. 
In such settings, sensors cover the same region of the scene, and each action instance is fully visible across all views.
Datasets such as Nanyang Technological University (NTU) RGB+D~\cite{shahroudy2016ntu, liu2019ntu} and NorthWestern-University of California at Los Angeles (NW-UCLA) Multiview Action 3D~\cite{wang2014cross} adopt this assumption, enabling methods~\cite{das2023viewclr, liu2023dual, shah2023multi, yang2026trunk} to exploit contrastive or disentanglement learning to enforce viewpoint invariance.  
In contrast, partially overlapping settings are more representative of real-world deployments.
These setups feature non-aligned sensor coverage, where an action may be visible in some views but completely occluded or out of scope in others.
Recent methods~\cite{yasuda2022multi, yasuda2024guided, nguyen2024action, nguyen2025multisensor} address this challenge by proposing fusion strategies to integrate fragmented evidences or recover missing observations.
However, they focus primarily on feature and view fusion, without explicitly enforcing consistency across views when actions are only partially observed. 
Therefore, their effectiveness in realistic partially overlapping scenarios is limited, and explicit modeling of view-aware consistency remains underexplored.


On the other hand, multi-view action recognition models achieve their best performance when all modalities are fully available and dense frame-level annotations are provided~\cite{nguyen2025multisensor}.  
However, in real-world multi-sensor deployments, these conditions are often unmet.
First, modality-limited scenarios occur when some input streams (e.g., audio or depth) are unavailable due to hardware or environmental constraints~\cite{10.1145/3711860}. 
Second, annotation-limited scenarios arise as frame-level labeling is prohibitively expensive for multi-view videos, leaving only weak sequence-level tags in many datasets~\cite{yasuda2022multi, yasuda2024guided}.
Although prior works~\cite{john2024frame, nguyen2024action, 10.1145/3744742} address learning under limited conditions, they primarily rely on weak supervision, remaining sensitive to fragmented observations and noisy labels.

To handle this limitation, the potential direction is to transfer supervision from a stronger model to a weaker one.
Knowledge Distillation (KD)~\cite{hinton2015distilling} offers a promising solution by allowing a fully supervised multi-modal teacher trained with strong labels to guide a modality- and annotation-limited student. 
Multi-modal KD (MKD)~\cite{gupta2016cross, garcia2021distillation, chen2021distilling, perez2020audio} extends this concept to cases where the teacher benefits from multi-modal inputs and frame-level labels, while the student operates with incomplete modalities or only coarse sequence-level labels.
However, applying KD to partially overlapping multi-view action recognition remains challenging, as it requires transferring knowledge while preserving consistency across views where the action may appear in some views but be absent in others.

To address these challenges, we propose \textbf{Vi}ew-aware \textbf{C}r\textbf{o}ss-modal \textbf{K}nowledge \textbf{D}istillation (ViCoKD), a method that distills knowledge from a fully supervised multi-modal teacher to a modality- and annotation-limited student. 
ViCoKD employs feature-level and logit-level KD, leveraging a cross-modal adapter that enables the student to exploit audio-visual correlations through Cross-modal Attention while operating solely on visual input.
In addition, we introduce a View-aware Consistency module that explicitly handles partially overlapping sensor setups by encouraging the model to produce the same prediction across different views whenever the action is co-visible.
This is achieved through a confidence-weighted Jensen–Shannon divergence~\cite{menendez1997jensen} between the predicted class distributions of co-visible views, ensuring that supervision is concentrated on frames with reliable, view-consistent evidence.
The main contributions of this study are as follows:  
\begin{itemize}
    \item \textbf{ViCoKD method:} A cross-modal attention-based KD method that transfers knowledge from a fully supervised multi-modal teacher to a modality- and annotation-limited student for multi-view action recognition in partially overlapping sensor settings.     
    \item \textbf{View-aware Consistency module:} A consistency module that leverages human-detection masks and confidence-weighted Jensen–Shannon divergence~\cite{menendez1997jensen} to align predictions only on view pairs with reliable and visible action evidence.
    \item \textbf{Comprehensive evaluation:} Experiments on the real-world MultiSensor-Home dataset~\cite{nguyen2025multisensor} across diverse backbones and home environments show that ViCoKD consistently outperforms competitive distillation methods and surpasses the teacher under limited conditions.
\end{itemize}

\noindent The remainder of this paper is structured as follows: Section~\ref{sec:realted_work} reviews related work. Section~\ref{sec:method} introduces the proposed ViCoKD. Section~\ref{sec:experimental_results} presents the experimental results and analysis. Finally, Section~\ref{sec:conclusion} concludes the paper.

\section{Related Work}
\label{sec:realted_work}


\noindent \textbf{Multi-view Action Recognition.}
Depending on sensor coverage, multi-view action recognition tasks are studied under either fully overlapping or partially overlapping settings.
Most prior work focuses on the former, with datasets such as NTU RGB+D~\cite{shahroudy2016ntu, liu2019ntu}, NW-UCLA~\cite{wang2014cross}, and Toyota Smarthome~\cite{das2019toyota} where all views capture the same subject. 
Building on these datasets, ViewCLR~\cite{das2023viewclr} employs view generation to generalize to unseen viewpoints, ViewCon~\cite{shah2023multi} applies supervised contrastive learning to produce viewpoint-invariant embeddings, and Dual-Recommendation Disentanglement Network (DRDN)~\cite{liu2023dual} leverages disentanglement learning to decouple action-specific and view-specific cues.
In contrast, real-world applications involve wide-area surveillance with only partially overlapping views, where actions may be visible from some sensors but entirely occluded in others. 
Recent work has introduced datasets tailored to this challenging setting~\cite{yasuda2024guided, nguyen2025multisensor}. 
Yasuda et al.~\cite{yasuda2022multi} introduce MultiTrans, which models inter-sensor relationships to integrate multi-view cues, and later propose Guided Masked sELf-Distillation (Guided-MELD) \cite{yasuda2024guided} to handle fragmented observations by reducing redundancy and filling missing sensor data for coherent event-level representations.
John and Kawanishi~\cite{john2024frame} present a weakly supervised latent embedding framework that learns from sequence-level labels while enabling frame-level action detection.
Recently, Nguyen et al.~\cite{nguyen2024action, nguyen2025multisensor} proposed MultiASL for weakly supervised action selection across sensors and MultiTSF for robust cross-sensor temporal fusion.
Despite these advances, most existing work focuses on feature and view fusion strategies, while explicit modeling of view-aware consistency in partially overlapping settings is still limited.

\begin{figure*}[t]
    \centering
    \begin{subfigure}[t]{1.0\textwidth}
        \centering
        \includegraphics[width=0.8\textwidth]{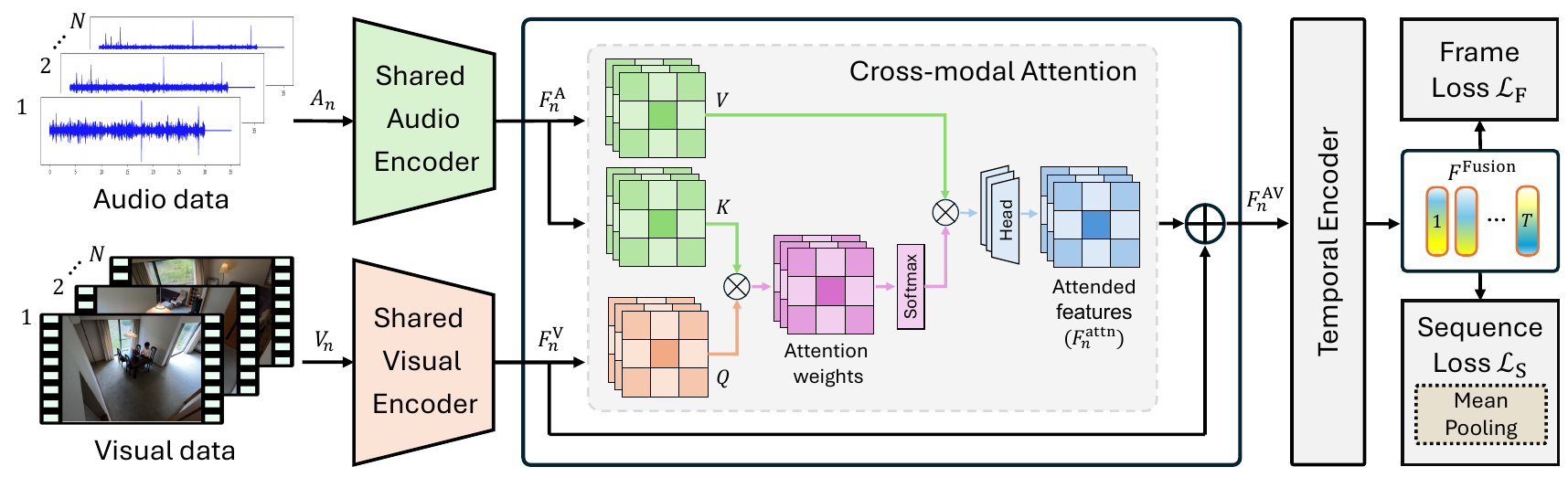}
        \caption{Multi-modal multi-view teacher network with cross-modal attention, trained with frame-level and sequence-level supervisions.}
        \label{fig:teacher_modal}
    \end{subfigure} %
    \hfill
    \begin{subfigure}[t]{1.0\textwidth}
        \centering
        \includegraphics[width=0.90\textwidth]{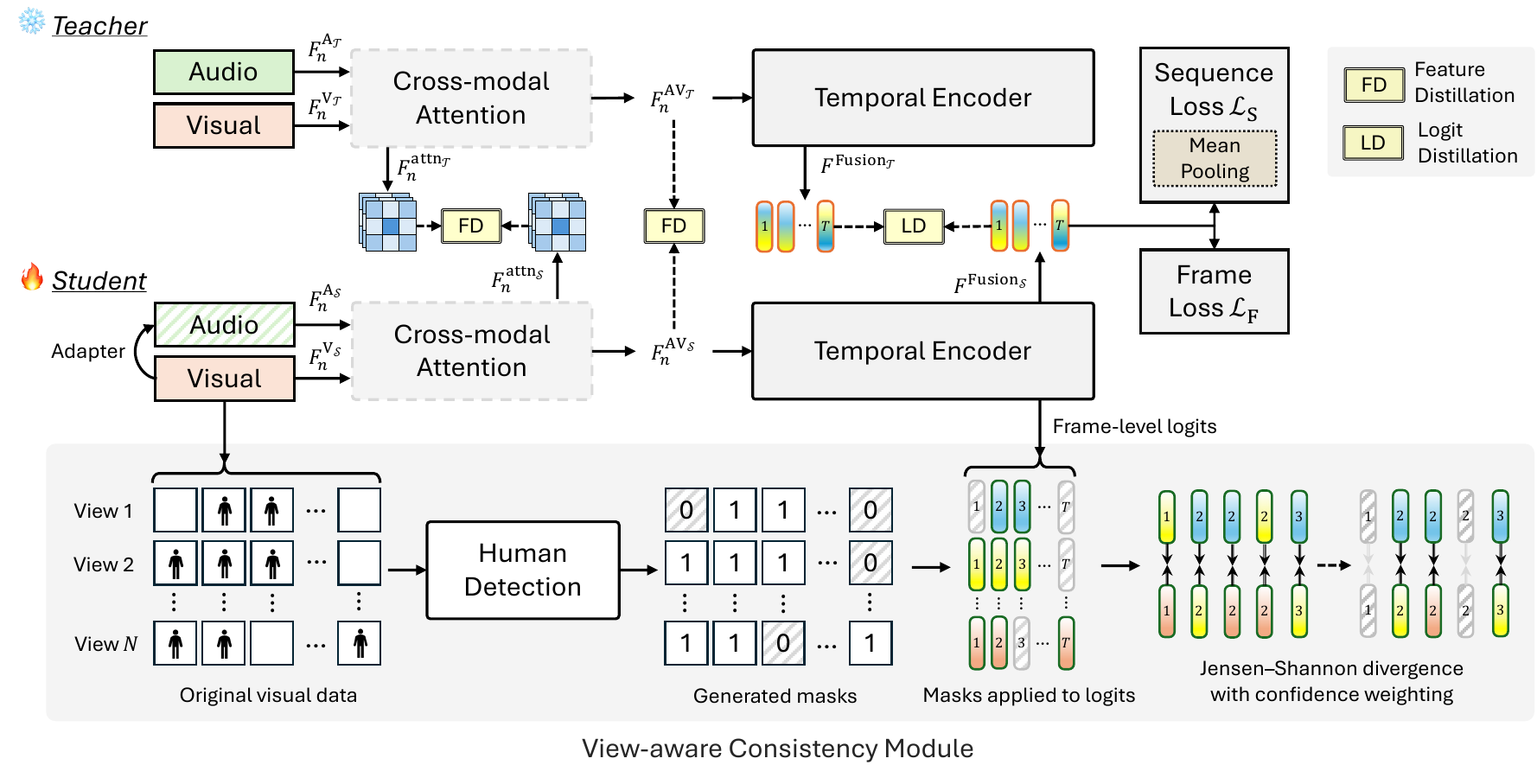}
        \caption{Knowledge distillation pipeline with a frozen teacher and a student trained via feature and logit distillation. 
        A cross-modal adapter generates pseudo-audio features when audio is unavailable, enabling cross-modal attention. 
        The view-aware consistency module applies human-detection-based masks and Jensen–Shannon divergence with confidence weighting for view-consistent supervision.
        }
        \label{fig:knowledge_distillation}
    \end{subfigure}
    \caption{
    Overview of the proposed ViCoKD method: (a) A multi-modal multi-view teacher with cross-modal attention, and (b) a knowledge distillation pipeline where the student trained using feature-level and logit-level distillation under view-aware consistency supervision.
    }
    \label{fig:overview}
    \vspace{-7pt}
\end{figure*}

\vspace{5pt}
\noindent \textbf{Multi-modal Knowledge Distillation (MKD).}
KD~\cite{hinton2015distilling} trains a smaller or weaker {student} model to mimic the predictions or intermediate representations of a teacher model, improving efficiency or performance in limited conditions. 
Beyond matching final logits~\cite{hinton2015distilling}, KD has evolved to include feature-level supervision~\cite{romero2015fitnetshintsdeepnets, zagoruyko2016paying} and relational knowledge transfer~\cite{park2019relational} to strengthen teacher--student alignment. 
MKD~\cite{chen2021distilling, zhang2021knowledge} extends KD to settings where the teacher and student operate over multiple modalities. 
The teacher typically leverages full-modality inputs, such as RGB, audio, depth, or skeleton data, while the student learns to replicate its performance under restricted modalities. 
Prior works have demonstrated the benefits of MKD for RGB-depth~\cite{gupta2016cross}, RGB-flow~\cite{garcia2021distillation}, and audio-visual~\cite{chen2021distilling, perez2020audio} tasks. 
In video understanding, MKD has been shown to improve recognition under missing modalities~\cite{radevski2023multimodal} by transferring temporal and cross-modal cues from the teacher to the student.
In this work, we study MKD in the context of partially overlapping multi-view action recognition under missing-audio conditions. 
Unlike these methods, our student operates solely with visual inputs and is trained under frame-level or sequence-level supervision, the latter representing a more challenging weakly supervised setting. 
To address this gap, we propose the \mbox{ViCoKD} method with View-aware Consistency, enabling students to learn robust cross-modal and cross-view representations under the limited conditions. 

\section{ViCoKD: View-aware Cross-modal Distillation Framework}
\label{sec:method}


We address the task of multi-view action recognition in partially overlapping settings within a KD framework.
Actions are captured from $N$ synchronized viewpoints. 
In this study, we address the multi-modal setting as comprising audio and visual inputs.
At each time step $t \in \{1, \dots, T\}$, the input from view $n \in \{1, \dots, N\}$ of teacher comprises two modalities: audio segment $a_t^n \in \mathbb{R}^F$, represented as a time-aligned spectrogram with $F$ frequency bins per frame, and visual frame $v_t^n \in \mathbb{R}^{H \times W \times C}$, where $H$ and $W$ denote spatial resolution, and $C$ is the number of color channels. 
Audio and visual inputs from view $n$ are denoted as $A^n = \{a_t^n\}_{t=1}^T$ and $V^n = \{v_t^n\}_{t=1}^T$, respectively.
The objective is to train a student model for multi-label action classification by transferring knowledge from a fully supervised multi-modal teacher model. 
Here, we focus on the audio-missing setting, where the student lacks access to the audio modality. 
Two supervision scenarios are considered for the student: (1) Frame-level supervision, where the student is trained with frame-level and sequence-level labels, and (2) Sequence-level supervision, where the student is trained using only video-level labels.
The goal is to exploit the feature representations learned by the teacher (Figure~\ref{fig:teacher_modal}) and distill this knowledge into a student model (Figure~\ref{fig:knowledge_distillation}) under missing-modality conditions.

\subsection{Teacher Model with Cross-modal Attention}
Figure~\ref{fig:teacher_modal} illustrates the teacher model, which is designed to learn spatio-temporal representations from synchronized multi-modal multi-view inputs, trained with strong supervision at the frame and sequence levels.

\vspace{5pt}
\noindent \textbf{Modality-specific Encoders.}
Each input sequence consists of audio data $A_n$ and visual data $V_n$ from view $n$, which are processed by modality-specific encoders. 
Specifically, we use Audio Spectrogram Transformer (AST)~\cite{gong2021ast} as the audio encoder $E_\text{A}$ and Vision Transformer (ViT)~\cite{dosovitskiy2020image} as the visual encoder $E_\text{V}$ as:
\begin{equation}
F_n^\text{A} = E_\text{A}(A_n), \quad F_n^\text{V} = E_\text{V}(V_n),
\end{equation}
where $F_n^\text{A} \in \mathbb{R}^{T \times D_a}$ and $F_n^\text{V} \in \mathbb{R}^{T \times D_v}$ are temporally aligned audio and visual feature sequences, respectively.
The AST and ViT encoders share parameters across all views to ensure consistent representation learning.

\vspace{5pt}
\noindent \textbf{Cross-modal Attention Module.}
To integrate audio-visual information, we adopt a cross-modal attention mechanism~\cite{ye2019cross, wei2020multi} where audio features are used as keys and values and visual features serve as queries. 
Positional encodings are added to both modalities to retain temporal and spatial ordering before cross-modal attention is applied as:
\begin{equation}
F_n^{\text{attn}} = \text{CrossAttn}(Q=F_n^\text{V}, K=F_n^\text{A}, V=F_n^\text{A}).
\end{equation}
The attended features $F_n^{\text{attn}}$ are fused with the visual features via element-wise addition as:
\begin{equation}
F_n^\text{AV} = F_n^\text{V} + F_n^{\text{attn}}.
\end{equation}
This fusion strategy preserves the spatial-temporal structure of the visual representation while enhancing it with complementary audio information.

\vspace{5pt}
\noindent \textbf{Temporal Encoding.}
The fused representation $F_n^\text{AV} \in \mathbb{R}^{T \times D}$ from each view $n$ is fed into a Transformer-based~\cite{vaswani2017attention} temporal encoder to capture sequential dependencies across time as:
\begin{equation}
F_n^{\text{Fusion}} = \text{TemporalEncoder}(F_n^\text{AV}),
\end{equation}
where $\text{TemporalEncoder}$ is a multi-layer Transformer that models self-attention across frames to learn long-range temporal patterns within the sequence. 
Output $F_n^{\text{Fusion}} \in \mathbb{R}^{T \times D'}$ represents a temporally enriched feature sequence encoding spatio-temporal cues.

\vspace{5pt}
\noindent \textbf{Supervision Objective.}
The teacher is optimized with frame-level and sequence-level classification losses.
At the frame-level, we predict per-frame logits and compute a Two-way loss~\cite{kobayashi2023two} that combines sample-wise and class-wise objectives. 
At the sequence-level, we apply temporal average pooling followed by classification, using the same loss formulation. 
The overall training objective for the teacher is calculated as:
\begin{equation}
\mathcal{L} = \mathcal{L}_\text{F} + \lambda \mathcal{L}_\text{S},
\label{eq:supervision_objective}
\end{equation}
where $\mathcal{L}_\text{F}$ and $\mathcal{L}_\text{S}$ are the frame-level and sequence-level classification losses, respectively, and $\lambda$ is a balancing hyperparameter.

\subsection{Teacher-to-student View-aware Cross-modal Knowledge Distillation}
Figure~\ref{fig:knowledge_distillation} illustrates the knowledge distillation pipeline, where the student $(\mathcal{S})$ adopts the same architecture as the teacher $(\mathcal{T})$ while operating with a missing audio modality.
The student is trained to mimic the teacher's knowledge through the supervision objective in Eq.~\eqref{eq:supervision_objective}, combined with a KD loss that transfers feature- and logit-level representations from the teacher. 
In addition, the student benefits from the View-aware Consistency loss, which enforces consistent predictions across partially overlapping views when the action is co-visible.

\vspace{5pt}
\noindent \textbf{Cross-modal Adapter.}
When audio is missing, a cross-modal adapter $\mathcal{A}_\psi$ is employed to synthesize pseudo-audio features $\hat{F}_n^\text{A} = \mathcal{A}_\psi(F_n^\text{V})$ from visual features $F_n^\text{V}$, where $\mathcal{A}_\psi$ is a lightweight feed-forward network with a non-linear activation.
The pseudo-audio feature is then used in cross-modal attention as:
\begin{equation}
F_n^{\text{attn}_S} = \text{CrossAttn}(Q=F_n^\text{V}, K=\hat{F}_n^\text{A}, V=\hat{F}_n^\text{A}).
\end{equation}

\vspace{5pt}
\noindent \textbf{Feature Distillation (FD).} 
We distill knowledge at two stages: (1) The attended audio-visual features $F_{n,t}^{\text{attn}}$ from the cross-modal attention, which transfer audio cues into the visual stream, and (2) The fused representations $F_{n,t}^\text{AV}$, which integrate multi-modal context.
Formally, we minimize the frame-wise Euclidean distance between the teacher's and the student's features as:
\begin{equation}
\mathcal{L}_{\text{FD}} = \frac{1}{T} \sum_{t=1}^{T} \left( 
\left\| F_{n,t}^{\text{attn}_\mathcal{T}} - F_{n,t}^{\text{attn}_\mathcal{S}} \right\|_2^2 +
\left\| F_{n,t}^{\text{AV}_\mathcal{T}} - F_{n,t}^{\text{AV}_\mathcal{S}} \right\|_2^2 \right).
\end{equation}
This alignment enhances the student's ability to inherit the teacher's cross-modal attention behavior and multi-modal feature integration.

\vspace{5pt}
\noindent \textbf{Logit Distillation (LD).} 
We apply logit-level distillation at the frame-level on the fused features $F^{\text{Fusion}}$ from multiple views. 
Let $p_t^\mathcal{T}$ and $p_t^\mathcal{S}$ denote the temperature-scaled softmax outputs of the teacher and student, respectively. 
The loss is computed as the Kullback--Leibler (KL) divergence~\cite{kullback1951information} between these distributions as:
\begin{equation}
\mathcal{L}_{\mathrm{LD}} = \frac{\tau^2}{T} \sum_{t=1}^T 
\mathrm{KL}\!\left(p_t^\mathcal{T} \,\|\, p_t^\mathcal{S}\right),
\end{equation}
where $\tau$ is the temperature parameter.

\vspace{5pt}
\noindent \textbf{View-aware Consistency Module.}
To enforce consistent predictions across views, we introduce a view-aware consistency loss applied at the frame-level.  
Figure~\ref{fig:knowledge_distillation} shows the details: For each view $n$ at time step $t$, we obtain probabilities $p_{n,t} \in \mathbb{R}^C$.  
A binary visibility mask $m_{n,t} \in \{0,1\}$ is generated using human detection~\cite{wang2024yolov10}, where $m_{n,t} = 1$ indicates that a human is visible in view $n$ at frame $t$, suggesting a high potential for action occurrence.  
Given a view pair $(i, j)$, we compute the Jensen--Shannon (JS) divergence~\cite{menendez1997jensen} between their prediction distributions as:
\begin{equation}
\mathrm{JS}(p_i, p_j) = \frac{1}{2} \mathrm{KL}\!\left(p_i \,\middle\|\, m\right) + \frac{1}{2} \mathrm{KL}\!\left(p_j \,\middle\|\, m\right),
\end{equation}
where $m = \frac{1}{2}(p_i + p_j)$.
The divergence is computed only on frames where $m_{i,t} = m_{j,t} = 1$, and is weighted by the product of the maximum class probabilities from both views to emphasize confident predictions.  
The View-aware Consistency loss is defined as:
\begin{equation}
\mathcal{L}_{\mathrm{VC}} = \frac{1}{N_{\mathrm{pairs}}T} \sum_{(i,j)} \sum_{t=1}^T w_{t}^{(i,j)} \, m_{i,t} m_{j,t} \, \mathrm{JS}\!\left(p_{i,t}, p_{j,t}\right),
\end{equation}
where $w_{t}^{(i,j)} = \max_{c} p_{i,t,c} \cdot \max_{c} p_{j,t,c}$ serves as a confidence weight, encouraging consistency when both views are individually certain in their predictions, and $N_{\mathrm{pairs}}$ is the number of view pairs.
This loss aligns predictions where the action is visible in both views and places greater emphasis on high-confidence predictions.

\section{Evaluation}
\label{sec:experimental_results}

\begin{figure}[t]
    \centering
    \includegraphics[width=0.465\textwidth]{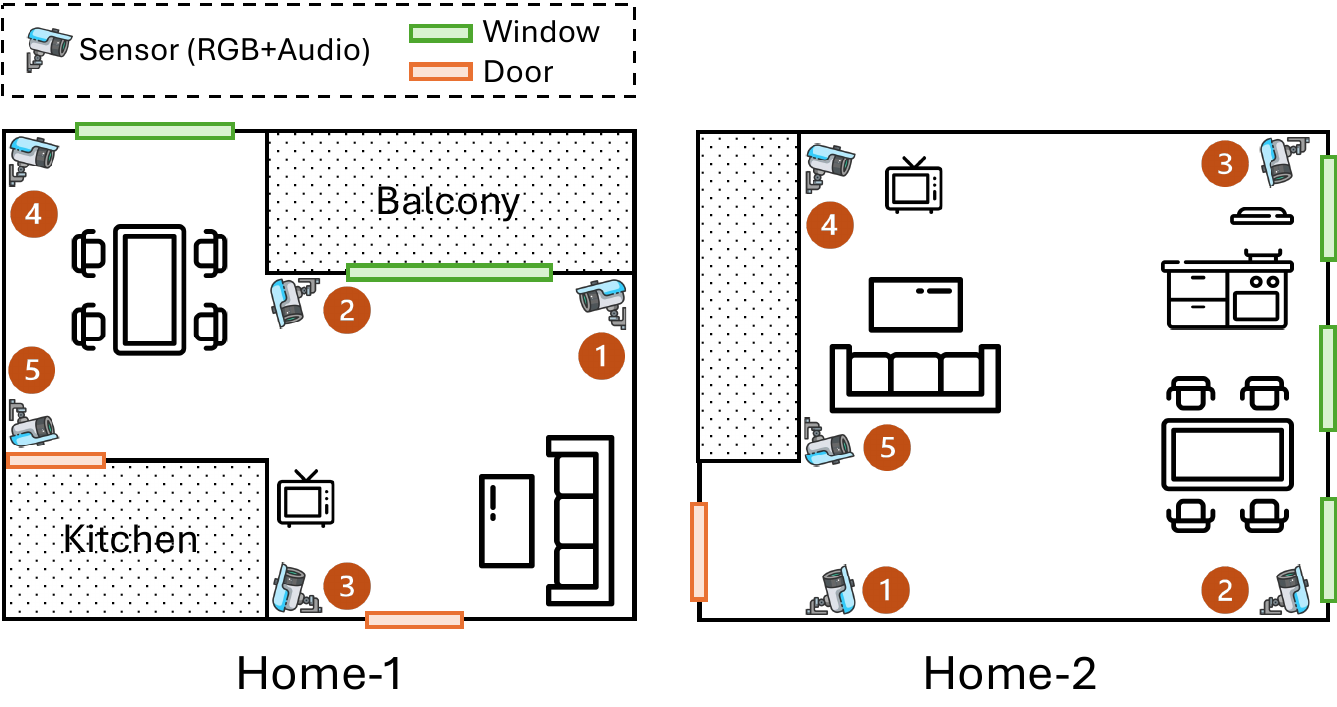}
    \caption{
    Room layouts and sensor views for the MultiSensor-Home dataset~\cite{nguyen2025multisensor} used in the experiments. 
    Each home environment is equipped with multiple RGB and Audio sensors, capturing scenes from different viewpoints with partial overlaps. 
    }
    \label{fig:dataset_setting}
    \vspace{-10pt}
\end{figure}

\subsection{Experimental Conditions}
\noindent \textbf{Data Preparation.}
We conduct the experiments on the publicly available MultiSensor-Home dataset~\cite{nguyen2025multisensor}, which includes two home environments: Home-1 and Home-2, as shown in Figure~\ref{fig:dataset_setting}. 
Following the official experimental setup, we use a 7:3 train/test split within each environment. 
To ensure a balanced class distribution, the iterative stratification strategy proposed by Sechidis et al.~\cite{sechidis2011stratification} is employed.
For all experiments, we extract a fixed number of \(T\) synchronized frames per sample. 
During training, we apply uniform sampling with slight random perturbations to generate frame sequences that cover the entire video while preserving the fixed length \(T\). 
This sampling strategy also augments temporal data, enhancing robustness by introducing sequence variability. 
For evaluation, we adopt deterministic uniform sampling without perturbation to ensure reproducibility across runs.


\vspace{5pt}
\noindent \textbf{Evaluation Metrics.}
Following~\cite{kobayashi2023two, nguyen2024action, nguyen2025multisensor}, we evaluate multi-label action recognition performance using mean Average Precision (mAP), a macro-averaged metric widely adopted in multi-label classification tasks.

\begin{table*}[t]
\centering
\caption{Comparison of the proposed ViCoKD method with baseline distillation methods in terms of mAP [\%].
Here, AV$_\text{F} \rightarrow$ V$_\text{F}$ and AV$_\text{F} \rightarrow$ V$_\text{S}$ denote distillation from an audio-visual teacher to a visual-only student with frame-level and sequence-level supervisions, respectively.
Numbers in parentheses show the difference from the non-distilled student baseline.
The best is emphasized using \best{bold}.
}
\setlength\tabcolsep{5pt} 
\resizebox{\textwidth}{!}
{
\begin{tabular}{l|l|cc|cc|cc}
    \toprule
    \multirowcell{2}[-2pt][l]{Environment} & \multirowcell{2}[-2pt][l]{Method} 
    & \multicolumn{2}{c|}{{MultiTrans}~\cite{yasuda2022multi}} 
    & \multicolumn{2}{c|}{MultiASL~\cite{nguyen2024action}} & \multicolumn{2}{c}{MultiTSF~\cite{nguyen2025multisensor}} \\
    \cmidrule(lr){3-4} \cmidrule(lr){5-6} \cmidrule(lr){7-8}
    & & AV$_\text{F}$ $\rightarrow$ V$_\text{F}$ & AV$_\text{F}$ $\rightarrow$ V$_\text{S}$ & AV$_\text{F}$ $\rightarrow$ V$_\text{F}$ & AV$_\text{F}$ $\rightarrow$ V$_\text{S}$ 
    & AV$_\text{F}$ $\rightarrow$ V$_\text{F}$ & AV$_\text{F}$ $\rightarrow$ V$_\text{S}$ 
    \\
    \midrule
    \multirow[t]{5}{*}{Home-1}
    & Teacher       & \multicolumn{2}{c|}{61.40} &  \multicolumn{2}{c|}{73.81}  & \multicolumn{2}{c}{76.12}   \\
    & Student       & 60.77 & 57.59 & 63.24 & 55.91 & 73.17 & 61.17 \\ \cmidrule(lr){2-8}
    & LogitKD       & \second{61.28} \scriptsize $(+0.51)$ & \second{60.48} \scriptsize $(+2.89)$ &  65.26 \scriptsize{$(\hspace{4pt}+2.02)$}  & 58.78 \scriptsize{$(+2.87)$} & 70.08 \scriptsize $(-3.09)$ & 60.67 \scriptsize $(\hspace{4pt}-0.50)$ \\
    & ModalKD       & 58.22 \scriptsize $(-2.55)$ & 54.04 \scriptsize $(-3.55)$ &  \second{69.50} \scriptsize $(\hspace{4pt}+6.26)$ & 58.92 \scriptsize $(+3.01)$ & \second{73.67} \scriptsize $(+0.50)$ & 62.88 \scriptsize $(\hspace{4pt}+1.71)$  \\
    & DualKD        & 58.84 \scriptsize $(-1.93)$ & 55.57 \scriptsize $(-2.02)$ &  68.42 \scriptsize $(\hspace{4pt}+5.18)$ & \second{60.45} \scriptsize $(+4.54)$ & 73.12 \scriptsize $(-0.05)$  & \second{63.22} \scriptsize $(\hspace{4pt}+2.05)$ \\
    & CoKD  & {61.34} \scriptsize $(+0.57)$ & {60.65} \scriptsize $(+3.06)$ & {75.33} \scriptsize{$(+12.09)$} & {64.52} \scriptsize{$(+8.61)$} & {76.89} \scriptsize $(+3.72)$ & {63.27} \scriptsize $(\hspace{4pt}+2.10)$ \\
    & ViCoKD (Ours)   & \best{66.27} \scriptsize $(+5.50)$ & \best{62.66} \scriptsize $(+5.07)$ & \best{76.83} \scriptsize $(+13.59)$ & \best{65.23} \scriptsize $(+9.32)$ & \best{82.91} \scriptsize$(+9.74)$ & \best{65.28} \scriptsize$(\hspace{4pt}+4.11)$ \\
    \cmidrule(lr){1-8}
    \multirow[t]{5}{*}{Home-2}
    & Teacher       &  \multicolumn{2}{c|}{86.60} & \multicolumn{2}{c|}{90.14} & \multicolumn{2}{c}{92.12} \\
    & Student       & 84.06 & 80.83 & 82.25 & 81.67 & 86.89 & 79.16 \\ \cmidrule(lr){2-8}
    & LogitKD       & 86.70 \scriptsize $(+2.64)$ & 85.55 \scriptsize $(+4.72)$ & 87.59 \scriptsize{$(\hspace{4pt}+5.34)$} & 84.94 \scriptsize{$(+3.27)$} & \second{88.60} \scriptsize $(+1.71)$ & 84.73 \scriptsize $(\hspace{4pt}+5.57)$ \\
    & ModalKD       & 85.43 \scriptsize $(+1.37)$ & 83.35 \scriptsize $(+2.52)$  & \second{91.45} \scriptsize $(\hspace{4pt}+9.20)$ & {87.98} \scriptsize $(+6.31)$ & {90.92} \scriptsize $(+4.03)$ & \second{87.56} \scriptsize $(\hspace{4pt}+8.40)$ \\
    & DualKD        & \second{86.41} \scriptsize $(+2.35)$ & \second{84.63} \scriptsize $(+3.80)$ & 85.58 \scriptsize $(\hspace{4pt}+3.33)$ & 81.88 \scriptsize $(+0.21)$ & 87.71 \scriptsize $(+0.82)$ & 85.36 \scriptsize $(\hspace{4pt}+6.20)$ \\ 
    & CoKD  & {88.64} \scriptsize $(+4.58)$ & {86.12} \scriptsize $(+5.29)$ & {91.77} \scriptsize{$(\hspace{4pt}+9.52)$} & \second{86.97} \scriptsize{$(+5.30)$} & 88.42 \scriptsize $(+1.53)$ & {87.75} \scriptsize $(\hspace{4pt}+8.59)$ \\
    & ViCoKD (Ours)   & \best{90.86} \scriptsize $(+6.80)$ & \best{89.31} \scriptsize $(+8.48)$ &  \best{91.80} \scriptsize $(\hspace{4pt}+9.55)$  & \best{89.89} \scriptsize $(+8.22)$ & \best{91.27} \scriptsize $(+4.38)$ & \best{89.21} \scriptsize $(+10.05)$  \\
    \bottomrule
\end{tabular}
}
\vspace{-5pt}
\label{table:main_table}
\end{table*}

\vspace{5pt}
\noindent \textbf{Comparison Methods.}
To evaluate the effectiveness of the proposed ViCoKD method, we compare against several representative KD methods under a consistent experimental setup. 
All methods adopt the same teacher--student configuration for a fair comparison.  
The teacher network is trained with frame-level audio-visual modalities (AV$_{\mathrm{F}}$), while the student is trained with visual-only inputs under either frame-level or sequence-level supervision (V$_{\mathrm{F}}$ or V$_{\mathrm{S}}$).
The teacher and student models use the architectures from the original MultiTrans~\cite{yasuda2022multi}, MultiASL~\cite{nguyen2024action}, and MultiTSF~\cite{nguyen2025multisensor} methods.
\begin{itemize}
    \item {Student/Teacher:}  
    The teacher is a full-modality model with RGB and audio modalities.  
    The student uses only the RGB modality, without KD, serving as the lower-bound baseline.

    \item LogitKD~\cite{thoker2019cross, radevski2023multimodal}: Logit-level KD methods, in which the student learns from the teacher's softened output logits using Kullback–Leibler divergence~\cite{hinton2015distilling}.
    \item ModalKD~\cite{liu2021semantics}: Feature-level KD method, in which the teacher network distills knowledge to the visual and pseudo-audio features of the student network.
    \item DualKD~\cite{garcia2018modality}: Feature-level and logit-level KD method, in which the teacher network distills knowledge to  features and logits of the student network.
    \item CoKD  (ViCoKD w/o Vi): The proposed ViCoKD method without the View-aware Consistency module.
\end{itemize}

\vspace{5pt}
\noindent \textbf{Models \& Hyperparameters.}
We adopt the backbone architecture and hyperparameters of the teacher and student networks from the original work.
Following Nguyen et al.~\cite{nguyen2025multisensor}, we sample input videos at 2.5~FPS and fix the sequence length to \(T = 70\) frames, based on the average video duration in the dataset.
The cross-modal adapter is implemented as a lightweight feed-forward module with two linear layers and a Rectified Linear Unit (ReLU)~\cite{agarap2018deep} activation in between. 
The human-detection masks are generated by a You Only Look Once (YOLO) v10-based~\cite{wang2024yolov10} detector.
All loss terms are equally weighted with coefficients set to \(1.0\) for simplicity. 
Optimization is performed using the Adaptive moment estimation (Adam)~\cite{kingma2014adam} optimizer with an initial learning rate of \(10^{-4}\), weight decay of \(5.0 \times 10^{-4}\), and a batch size of $12$ for $100$ training epochs. 
The learning rate is scheduled using cosine annealing~\cite{loshchilov2016sgdr} for gradual decay over the course of training.

\subsection{Quantitative Results}

Table~\ref{table:main_table} presents the mAP results of the proposed ViCoKD method compared with competitive KD baselines across three backbones: MultiTrans~\cite{yasuda2022multi}, MultiASL~\cite{nguyen2024action}, and MultiTSF~\cite{nguyen2025multisensor}.
The evaluation covers KD supervision scenarios: AV$_\text{F}$ $\rightarrow$ V$_\text{F}$ and AV$_\text{F}$ $\rightarrow$ V$_\text{S}$, which denote distillation from an audio-visual teacher to a visual-only student with frame-level and sequence-level supervisions, respectively.

\begin{figure*}[t]
    \centering
    \begin{subfigure}[t]{0.4965\textwidth}
        \centering
        \includegraphics[width=1.0\textwidth]{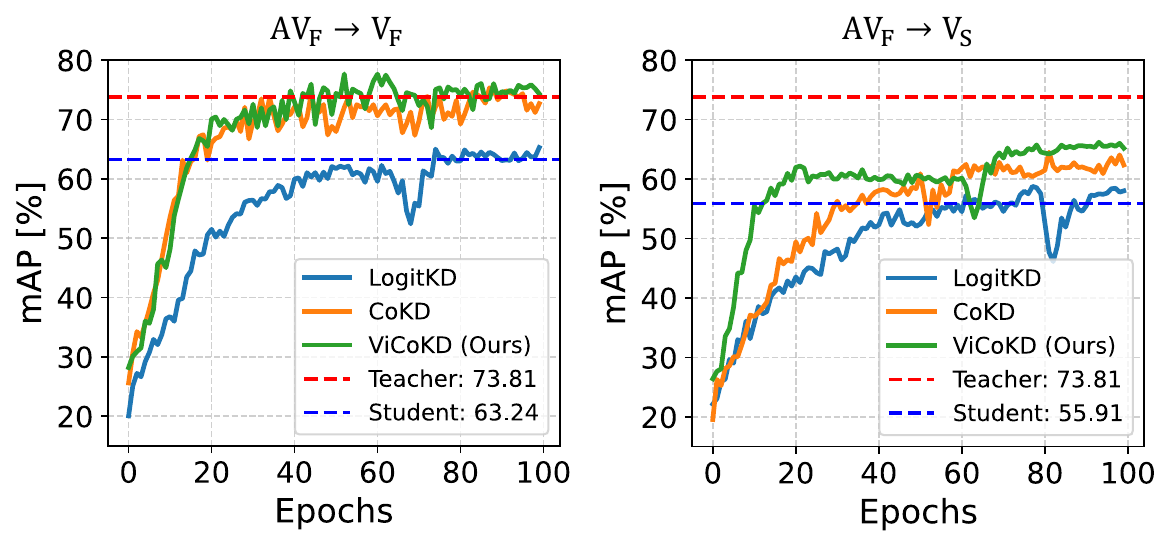}
        \caption{Home-1 environment.}
        \label{fig:test_curve_home1}
    \end{subfigure} %
    \hfill
    \begin{subfigure}[t]{0.490\textwidth}
        \centering
        \includegraphics[width=1.0\textwidth]{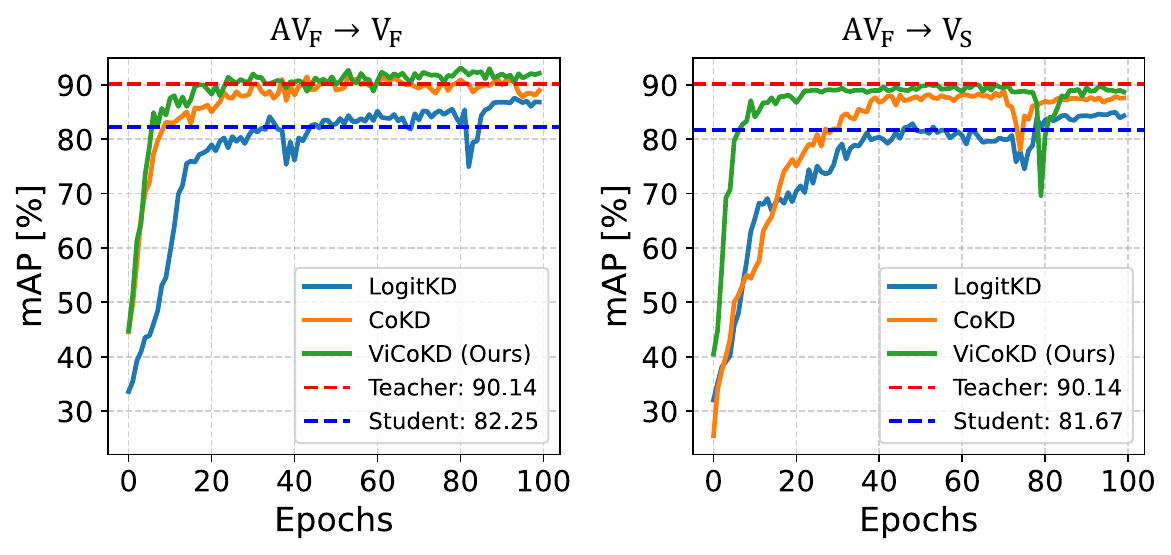}
        \caption{Home-2 environment.}
        \label{fig:test_curve_home2}
    \end{subfigure}
    \caption{mAP [\%] curves on the test set using the MultiASL~\cite{nguyen2024action} backbone under different distillation settings.
    }
    \label{fig:test_curve_MultiASL}
    \vspace{-10pt}
\end{figure*}

\vspace{5pt}
\noindent \textbf{Home-1 Environment.}
This environment is particularly challenging due to its greater view disparity and weaker inter-view correlation.
ViCoKD consistently delivered the highest mAP across all backbones and both KD supervision levels, with particularly large gains in the sequence-level regime.
Using the MultiASL~\cite{nguyen2024action} backbone, ViCoKD improved mAP over the non-distilled student by $+13.59$ (AV$_\text{F}$ $\rightarrow$ V$_\text{F}$) and $+9.32$ (AV$_\text{F}$ $\rightarrow$ V$_\text{S}$), surpassing the teacher in both cases.
The gains over CoKD (ViCoKD w/o Vi) across all experiments for both frame-level and sequence-level supervisions highlight the contribution of the proposed View-aware Consistency module in leveraging spatially localized cues under partial overlaps.
In contrast, methods such as ModalKD and DualKD often underperformed the teacher, while LogitKD occasionally performed worse than the non-distilled student, underscoring their inability to adapt to scenarios where cross-view correspondence is weak or sparse.

\vspace{5pt}
\noindent \textbf{Home-2 Environment.}
In this environment, the sensor views are closer in what they capture, which makes the setting less challenging than Home-1. 
Under these conditions, ViCoKD still secured top performance across all backbones.
It achieved mAP improvements of up to $+9.55$ (AV$_\text{F}$ $\rightarrow$ V$_\text{F}$) and $+10.05$ (AV$_\text{F}$ $\rightarrow$ V$_\text{S}$) over the non-distilled student baseline.
While LogitKD, ModalKD, and DualKD performed competitively in AV$_\text{F}$ $\rightarrow$ V$_\text{F}$, their advantages diminished under sequence-level supervision (AV$_\text{F}$ $\rightarrow$ V$_\text{S}$), reflecting the difficulty of exploiting weak labels.
The consistent margins over CoKD reaffirm the effectiveness of the View-aware Consistency module in integrating human-centric spatial constraints into partially overlapping setups.

\begin{table}[t]
\centering
\caption{Mean $\pm$ standard deviation of mAP [\%] over three runs for the proposed ViCoKD method using different backbones.}
\setlength\tabcolsep{3pt} 
\resizebox{\columnwidth}{!}
{
\begin{tabular}{l|cc|cc}
    \toprule
    \multirowcell{2}[-2pt][l]{Backbone}  & \multicolumn{2}{c|}{Home-1} & \multicolumn{2}{c}{Home-2} \\
    \cmidrule(lr){2-3} \cmidrule(lr){4-5}
           & AV$_\text{F}$ $\rightarrow$ V$_\text{F}$ & AV$_\text{F}$ $\rightarrow$ V$_\text{S}$ & AV$_\text{F}$ $\rightarrow$ V$_\text{F}$ & AV$_\text{F}$ $\rightarrow$ V$_\text{S}$ \\
    \midrule
    MultiTrans & 66.27 \scriptsize $\pm$ 0.59 & 62.66 \scriptsize $\pm$ 0.77 &  90.86 \scriptsize $\pm$ 0.36 & 89.31 \scriptsize $\pm$ 0.40 \\
    MultiASL &  76.83 \scriptsize $\pm$ 0.80  & 65.23 \scriptsize $\pm$ 0.74 &  91.80 \scriptsize $\pm$ 0.85 &  89.89 \scriptsize $\pm$ 0.57 \\
    MultiTSF &  82.91 \scriptsize $\pm$ 0.65 &  65.28 \scriptsize $\pm$ 0.21 &  91.27 \scriptsize $\pm$ 0.79  &  89.21 \scriptsize $\pm$ 0.70 \\
    \bottomrule
\end{tabular}
}
\label{table:mean_std}
\vspace{-10pt}
\end{table}

\vspace{5pt}
\noindent \textbf{Key Findings.}
Across environments and backbones, ViCoKD consistently attained the highest performance, with particularly large gains in sequence-level supervision (AV$_\text{F}$ $\rightarrow$ V$_\text{S}$), where label sparsity is most severe.
Moreover, as shown in Table~\ref{table:mean_std}, \mbox{ViCoKD} achieved high mAP while maintaining low variance ($\le 0.85$ of mAP) over three runs, indicating stable convergence and robustness.
Figure~\ref{fig:test_curve_MultiASL} further demonstrates that \mbox{ViCoKD} converged faster and reached higher mAP than LogitKD and CoKD across all supervision settings and environments.
The clearest margin was observed under AV$_\text{F}$ $\rightarrow$ V$_\text{S}$ in both environments, where ViCoKD surpassed the non-distilled student early in training and maintained high performance, while LogitKD and CoKD achieved lower scores or exhibit fluctuations.
In addition, Table~\ref{table:impact_view_aware_consistency} shows that applying the View-aware Consistency module to the teacher network itself yielded consistent mAP improvements. 
These results indicate the general effectiveness of the module as a principled mechanism for improving multi-view representation learning.

\begin{table}[t]
\centering
\caption{Impact of the View-aware Consistency module when applied to the teacher network. $a \rightarrow b$ denotes the change in mAP from the original teacher ($a$) to the teacher with view-aware consistency ($b$).} 
\resizebox{1.0\columnwidth}{!}
{
\begin{tabular}{l|c|c}
  \toprule
  \multirowcell{1}[0pt][l]{Backbone} & \multicolumn{1}{c|}{Home-1} & \multicolumn{1}{c}{Home-2} \\
  \midrule
  MultiTrans~\cite{yasuda2022multi}         & 61.40 $\rightarrow$ 66.31 \scriptsize $(+4.91)$  & 86.60 $\rightarrow$ 88.88 \scriptsize $(+2.28)$ \\
  MultiASL~\cite{nguyen2024action}          & 73.81  $\rightarrow$ 75.99 \scriptsize $(+2.18)$ & 90.14  $\rightarrow$ 91.78 \scriptsize $(+1.64)$ \\
  MultiTSF~\cite{nguyen2025multisensor}     & 76.12  $\rightarrow$ 78.25 \scriptsize $(+2.13)$ & 92.12  $\rightarrow$ 92.61 \scriptsize $(+0.49)$ \\
  \bottomrule
\end{tabular}
}
\label{table:impact_view_aware_consistency}
\vspace{-5pt}
\end{table}

\subsection{Ablation Studies}
We conducted ablation studies to quantify the contribution of each component in the proposed ViCoKD method, with the experimental results shown in Table~\ref{table:cmvd_frame_to_seq}.
Note that some changes resulted in differences smaller than $0.85$ of mAP, which can be considered minor and likely within run-to-run variation, while others caused substantial drops, as indicated in Table~\ref{table:mean_std}.

\vspace{5pt}
\noindent \textbf{View-aware Consistency Module.}  
Eliminating Confidence Weighting (w/o ConfW) or the human-detection mask (w/o Mask) yielded notable drops of up to $-4.50$ of mAP, particularly in the Home-1 environment within AV$_\text{F}$ $\rightarrow$ V$_\text{S}$.
This indicates that weighting supervision by prediction reliability and masking irrelevant views are both crucial for robust cross-view alignment.  
Replacing the divergence from JS to KL (JS $\to$ KL) also caused a small performance drop, showing the benefit of using a symmetric divergence.

\vspace{5pt}
\noindent \textbf{Knowledge Distillation Strategies.}  
Removing Feature-level Distillation (w/o FD) resulted in small drops in most settings, whereas omitting Logit-level Distillation (w/o LD) caused a large decrease in AV$_\text{F}$ $\rightarrow$ V$_\text{S}$.
This suggests that while both FD and LD contributed to the overall improvement, LD was particularly important in scenarios with only sequence-level supervision.

\begin{table}[t]
\centering
\caption{Ablation study on the proposed ViCoKD method. 
Results are reported in mAP [\%] using the MultiASL~\cite{nguyen2024action} backbone.
Numbers in parentheses show the difference from ViCoKD.
}
\setlength\tabcolsep{3pt} 
\resizebox{\columnwidth}{!}
{
\begin{tabular}{l|cc|cc}
    \toprule
    \multirowcell{2}[-2pt][l]{Method}  & \multicolumn{2}{c|}{Home-1} & \multicolumn{2}{c}{Home-2} \\
    \cmidrule(lr){2-3} \cmidrule(lr){4-5}
           & AV$_\text{F}$ $\rightarrow$ V$_\text{F}$ & AV$_\text{F}$ $\rightarrow$ V$_\text{S}$ & AV$_\text{F}$ $\rightarrow$ V$_\text{F}$ & AV$_\text{F}$ $\rightarrow$ V$_\text{S}$ \\
    \midrule
    ViCoKD & {76.83} & {65.23} & 91.80 & 89.89 \\ \cmidrule(lr){1-5}
     \multicolumn{5}{c}{\textit{View-aware Consistency Module}} \\
    \cmidrule(lr){1-5}
    ~w/o ConfW        & 73.15 \scriptsize $(-3.68)$ & 62.27 \scriptsize $(-2.96)$ & 90.51 \scriptsize $(-1.29)$ & 88.90 \scriptsize $(-0.99)$ \\
    ~w/o Mask         & 72.33 \scriptsize $(-4.50)$ & 63.34 \scriptsize $(-1.89)$ & 90.08 \scriptsize $(-1.72)$ & 89.19 \scriptsize $(-0.70)$ \\
    ~JS $\to$ KL      & 75.50 \scriptsize $(-1.33)$ & 64.87 \scriptsize $(-0.36)$ & 90.11 \scriptsize $(-1.69)$ & 89.62 \scriptsize $(-0.27)$ \\
    \cmidrule(lr){1-5}
    \multicolumn{5}{c}{\textit{Knowledge Distillation Strategies}} \\
    \cmidrule(lr){1-5}
    ~w/o FD        & 76.52 \scriptsize $(-0.31)$ & 64.85 \scriptsize $(-0.38)$ & 90.44 \scriptsize $(-1.36)$ & 88.90 \scriptsize $(-0.99)$ \\
    ~w/o LD        & 76.23 \scriptsize $(-0.60)$ & 61.51 \scriptsize $(-3.72)$  & 92.02 \scriptsize $(+0.22)$ & 88.76 \scriptsize $(-1.13)$ \\
    \bottomrule
\end{tabular}
}
\label{table:cmvd_frame_to_seq}
\vspace{-5pt}
\end{table}

\begin{figure*}[t]
    \centering
    \begin{subfigure}[t]{0.49\textwidth}
        \centering
        \includegraphics[width=1.0\textwidth]{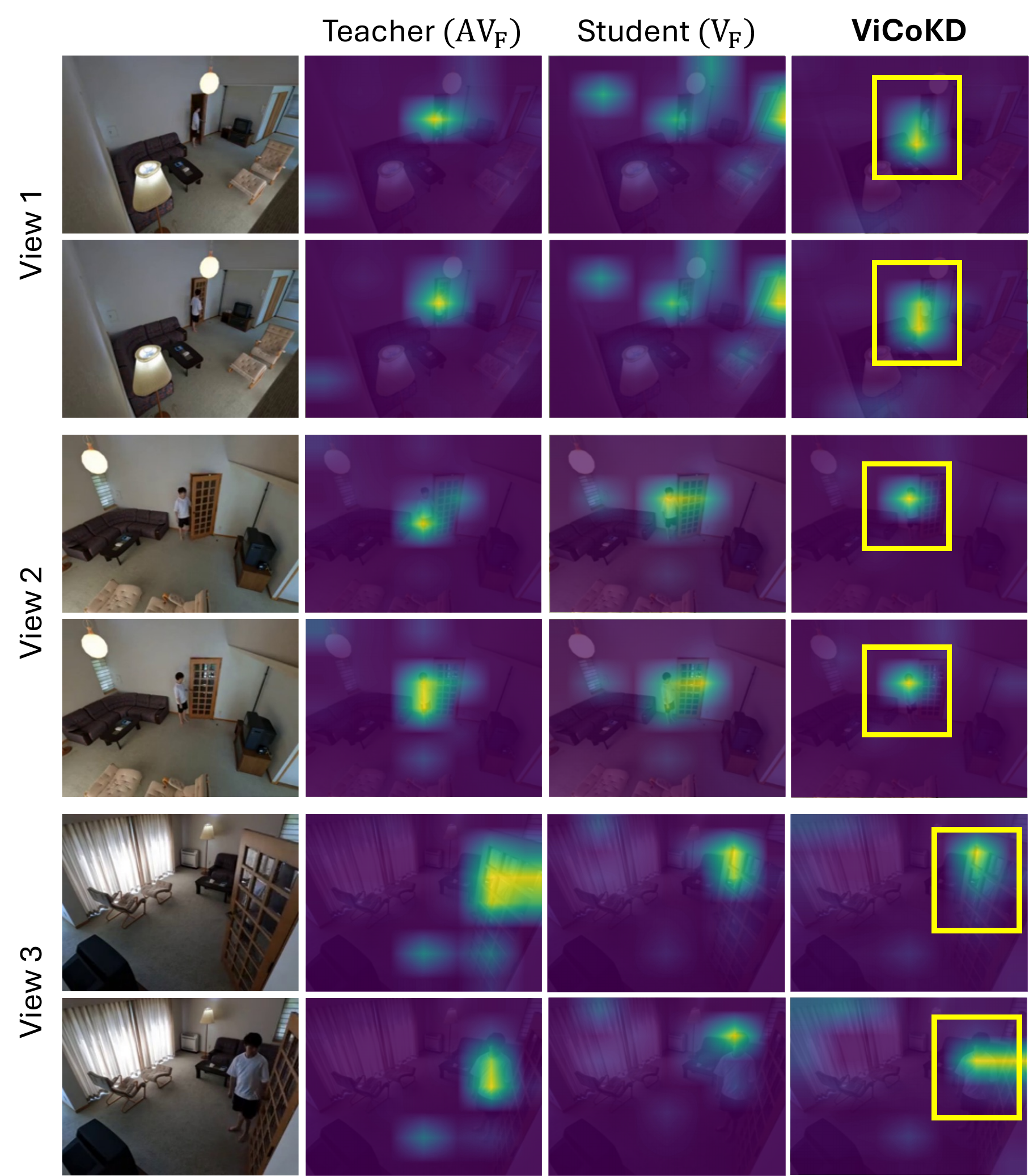}
        \caption{Home-1 environment: Action \textit{``enter the room''} on Views 1, 2, and 3.}
        \label{fig:attention_a}
    \end{subfigure} %
    \hfill
    \begin{subfigure}[t]{0.49\textwidth}
        \centering
        \includegraphics[width=\textwidth]{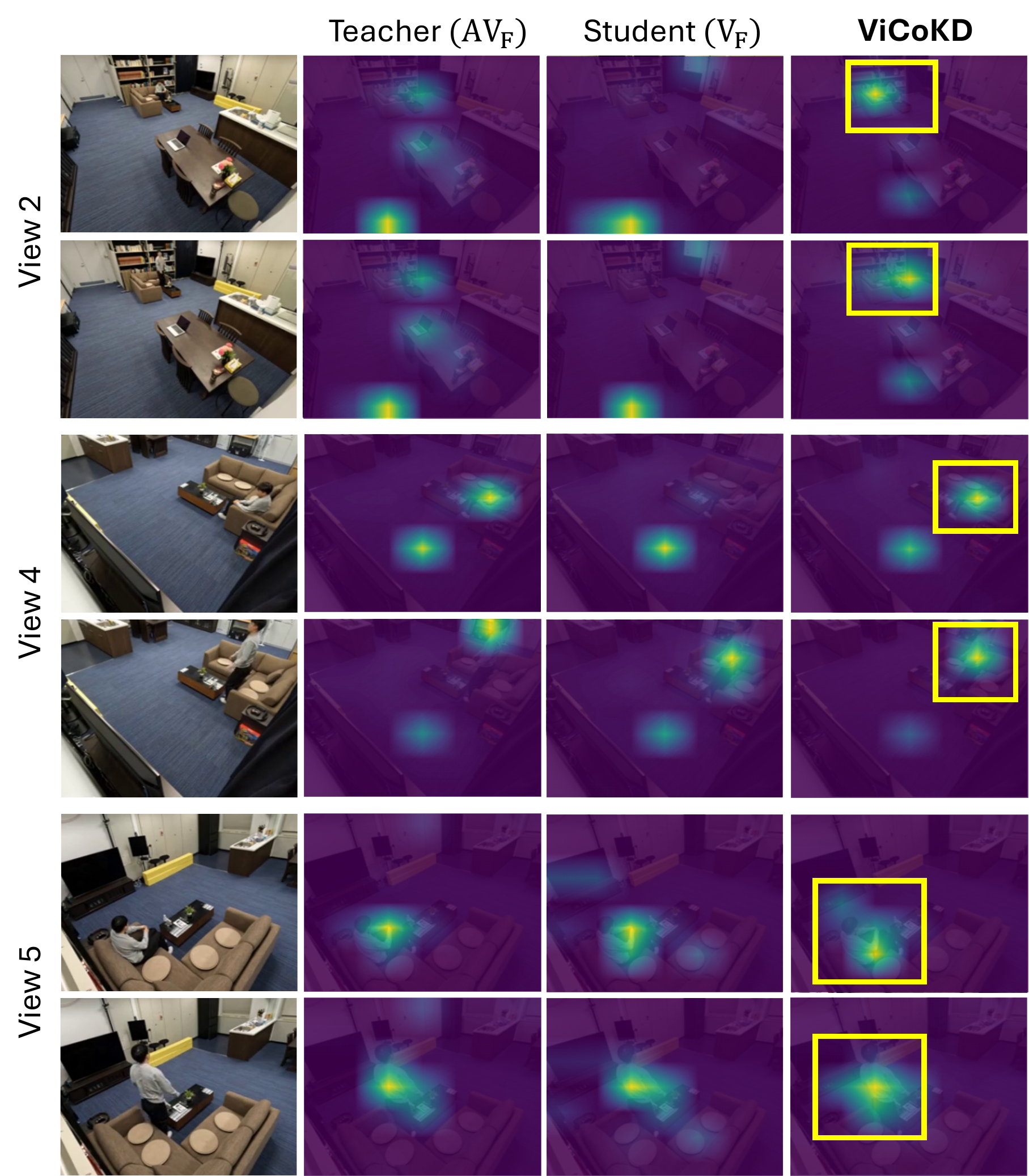}
        \caption{Home-2 environment: Action \textit{``stand up''} on Views 2, 4, and 5.}
        \label{fig:attention_b}
    \end{subfigure}
    \caption{Qualitative comparison of attention maps for the teacher, baseline student, and the proposed ViCoKD method using the MultiASL~\cite{nguyen2024action} backbone on the MultiSensor-Home dataset~\cite{nguyen2025multisensor}.  
    Each row corresponds to a different sensor view.  
    ViCoKD produces more precise and human-centric attentions (yellow boxes) compared to the baseline student. 
    }
    \label{fig:attention}
    \vspace{-5pt}
\end{figure*}

\subsection{Qualitative Results}
Figure~\ref{fig:attention} compares the attention maps of the teacher, the baseline non-distilled student, and the proposed ViCoKD for two representative actions across multiple sensor views.
In the Home-1 environment example (``enter the room''), the baseline student often exhibited diffuse or background-focused attention, missing the key human regions.
In contrast, ViCoKD consistently produced sharp, human-centric activation that tightly aligned with the actor's location across all views, even when the viewpoint changed substantially.
Similarly, in the Home-2 environment example (``stand up''), the attention of the student was scattered and shifted toward irrelevant areas such as furnitures. 
In contrast, ViCoKD focused precisely on the subject's body, capturing the motion cues critical for action recognition.
In several views, ViCoKD attended more precisely to human regions where the action occured, indicating that the proposed method transfered and refined the teacher's learned features, enabling the model to focus on discriminative regions despite partial view overlaps and reduced modalities.

\subsection{Limitations and Future Work}
\label{sec:limiation}
While the proposed ViCoKD method demonstrated strong performance in partially overlapping multi-view action recognition, it has several limitations.
First, the framework relies on pre-trained human detectors to generate human-detection masks, which may introduce errors under heavy occlusion, poor lighting, or unconventional poses. These errors can affect the reliability of View-aware Consistency supervision.
Second, the method depends on a strong multi-modal teacher, which limits applicability in domains where high-quality multi-modal data are scarce or unavailable.
Finally, the current experiments are limited by available multi-view action recognition datasets, which typically provide only audio and visual modalities and lack other informative sources such as depth or skeleton signals. 

\noindent Future work will explore end-to-end learned visibility estimation for improved robustness, as well as extending ViCoKD to additional modalities and cross-domain generalization.

\section{Conclusion}
\label{sec:conclusion}

We addressed the challenge of multi-view action recognition in partially overlapping sensor setups under modality and annotation-limited conditions. 
We introduced \textbf{Vi}ew-aware \textbf{C}r\textbf{o}ss-modal \textbf{K}nowledge \textbf{D}istillation (ViCoKD), a novel knowledge distillation method that transfers supervision from a fully supervised multi-modal teacher to a constrained student through cross-modal attention and a view-aware consistency mechanism. 
The proposed human-detection masks with confidence-weighted Jensen–Shannon divergence~\cite{menendez1997jensen} ensure that distillation focuses on frames with reliable, view-consistent evidence.
Extensive experiments on the MultiSensor-Home~\cite{nguyen2025multisensor} dataset demonstrated that ViCoKD achieved substantial gains over non-distilled students and competitive baselines, with consistent improvements across backbones and environments. 
These results highlight the importance of explicitly modeling View-aware Consistency in partially overlapping scenarios and provide a foundation for more robust multi-view action recognition in real-world deployments.

\section*{Acknowledgment}
This work was partly supported by JSPS KAKENHI JP21H03519 and JP24H00733. 
The computation was carried out using the General Projects on supercomputer ``Flow'' at IT Center, Nagoya University.

{
    \small
    \bibliographystyle{ieeenat_fullname}
    \bibliography{main}
}


\end{document}